\documentclass[mlabstract]{jmlr}

\usepackage{longtable}

\usepackage{booktabs}
\usepackage[load-configurations=version-1]{siunitx} 

\usepackage[capitalize]{cleveref}

\theorembodyfont{\upshape}
\theoremheaderfont{\scshape}
\theorempostheader{:}
\theoremsep{\newline}

\jmlrvolume{}
\firstpageno{1}
\editors{Christian Shewmake, Simone Azeglio, Bahareh Tolooshams, Sophia Sanborn, Nina Miolane}

\jmlryear{2024}
\jmlrworkshop{Symmetry and Geometry in Neural Representations}

\title[MNIST-Nd - datasets to benchmark clustering across dimensions]{MNIST-Nd: a set of naturalistic datasets to benchmark clustering across dimensions}
\author{
\Name{Polina Turishcheva\textsuperscript{1}} \Email{turishcheva@cs.uni-goettingen.de}\\
\Name{Laura Hansel\textsuperscript{1}} \Email{hansel@cs.uni-goettingen.de}\\ 
\Name{Martin Ritzert\textsuperscript{1}} \Email{ritzert@informatik.uni-goettingen.de}\\ 
\Name{Marissa A. Weis\textsuperscript{1}} \Email{marissa.weis@uni-goettingen.de}\\
\Name{Alexander S. Ecker\textsuperscript{1,2}} \Email{ecker@cs.uni-goettingen.de}\\
    \addr [1] Institute of Computer Science and Campus Institute Data Science, University of Göttingen, Germany
    \\\addr [2] Max Planck Institute for Dynamics and Self-Organization, Göttingen, Germany
}

% ...

\begin{document}

\maketitle

\begin{abstract}
Driven by advances in recording technology, large-scale high-dimensional datasets have emerged across many scientific disciplines.
Especially in biology, clustering is often used to gain insights into the structure of such datasets, for instance to understand the organization of different cell types. 
However, clustering is known to scale poorly to high dimensions, even though the exact impact of dimensionality is unclear as current benchmark datasets are mostly two-dimensional. 
Here we propose MNIST-Nd, a set of synthetic datasets that share a key property of real-world datasets, namely that individual samples are noisy and clusters do not perfectly separate. 
MNIST-Nd is obtained by training mixture variational autoencoders with 2 to 64 latent dimensions on MNIST, resulting in six datasets with comparable structure but varying dimensionality.
It thus offers the chance to disentangle the impact of dimensionality on clustering.
Preliminary common clustering algorithm benchmarks on MNIST-Nd suggest that Leiden is the most robust for growing dimensions.


\end{abstract}
\begin{keywords}
benchmarking datasets; clustering; high dimensional space
\end{keywords}

\section{Introduction}
\label{sec:intro}

Modern datasets are often high-dimensional, especially with deep learning embeddings \citep{schroff2015facenet, weis2022large, douze2024faiss} and advanced recording techniques in natural sciences, i.e. transcriptomics \citep{qiu2017reversed, wolf2019paga, harris2018classes, lause2024art}.
To uncover their internal structure, dimensionality reduction or clustering is commonly used.
While linear methods like PCA miss non-linear patterns, t-SNE \citep{van2008visualizing}, UMAP \citep{mcinnes2018umap} or PHATE \citep{Moon120378}, are popular for visualization but sensitive to hyperparameters \citep{kobak2019art, kobak2021initialization}, making conclusions based only on them challenging. 
Alternatively, clustering in the original space avoids information loss, but distinguishing distances or densities in high dimensions is difficult as pairwise distances become more alike 
\citep{johnstone2009statistical}. 

Many different clustering methods exist, but realistic benchmarks with non-uniform noise and high dimensional data are lacking.
Instead, most benchmarking datasets are 2D or 3D \citep{karypis2002cluto, gagolewski2022framework, thrun2020clustering, barton2015clustering, laborde2023sparse}.
Simple real-world datasets with high-dimensional features are also used for clustering evaluation \citep{thrun2020clustering}, but they are not directly comparable across dimensionalities as different datasets have different internal structures.
%
A few artificial datasets with variable dimensions exist, like multidimensional Gaussians \citep{gagolewski2022framework, laborde2023sparse, sedlmair2012taxonomy}
worms \citep{sieranoja2019fast}, and DENSIRED \citep{DENSIRED}. 
However, they lack realistic non-uniform noise and don’t scale variance with dimensionality, which makes cluster separation easier in high dimensions as clusters shrink and no longer overlap while overlapping density is common in real data.

To overcome these issues, we propose MNIST-Nd, a dataset of embeddings from a mixture variational autoencoder (m-VAE) \citep{m-vae}, trained on MNIST \citep{lecun1998mnist}.
MNIST-Nd has realistic noise as it appears in learned embeddings and controllable dimensions while maintaining consistent signal-to-noise ratio across dimensions. 
We benchmark common clustering algorithms ($k$-means, GMM, TMM, Leiden) on MNIST-Nd with respect to their performance and robustness across dimensions and find that Leiden clustering \citep{traag2019louvain} significantly outperforms other methods in higher dimensions for both performance and robustness.

\section{Dataset Creation}
\label{sec:method}
\begin{figure}[t]
  \includegraphics[width=\linewidth]{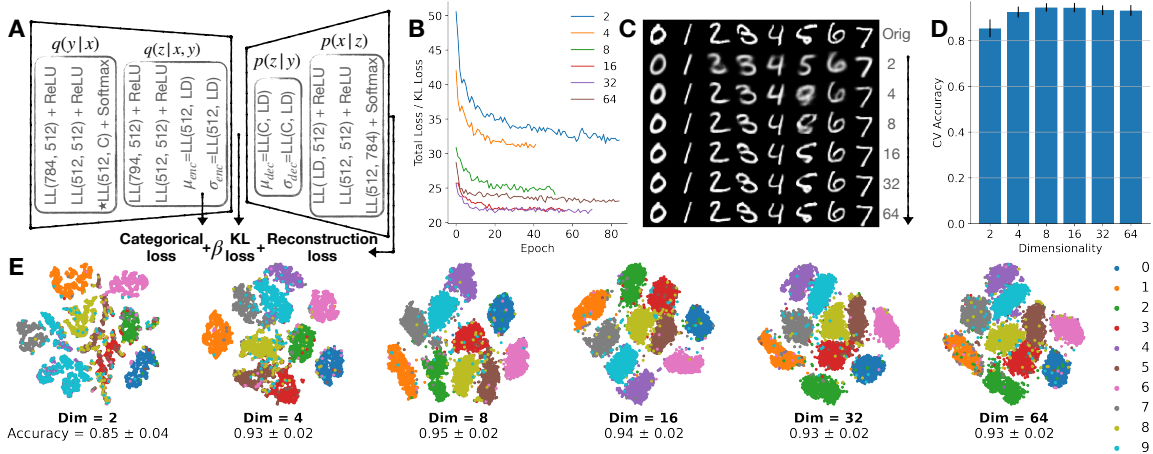}
  \caption{
  \textbf{A}: Mixture VAE architecture. LL: linear layer (+number of in and out channels). LD: latent dim. $\star$-layer: Gumbel softmax. 
  It returns a one-hot-encoding to select the mixture component. The next layer concatenates it to the original input (\cref{appendix:losses}).
  \textbf{B}: Total loss divided by KL-loss is similar across dimensions. KL: Kullback-Leibler Divergence. 
  \textbf{C}: Reconstruction examples. 
  \textbf{D}: Cross-validated random forest accuracy ($\pm$SD). 
  \textbf{E}:
  t-SNE visualizations of the embeddings.
  }
  \label{fig:vae1}
\end{figure}
To create the datasets, we use a mixture VAE \citep[m-VAE; \cref{fig:vae1}\,A;][]{m-vae}, where the prior is a mixture of Gaussians with ten components, which biases the latent space to have ten density modes. 
We use the $\beta$-VAE framework \citep{higgins2017beta} to scale the importance of the Kullback-Leibler (KL) loss.
As the KL loss is unbounded and grows with the number of dimensions,
we scale $\beta$ inversely proportional to the dimensionality such that all datasets are similarly regularized to match the prior shape.
The ratio of the total loss to the KL term converges to similar values across latent dimensions (\cref{fig:vae1}\,B), suggesting that the impact of the KL loss is indeed comparable.
The architecture of the autoencoder as well as all other training hyperparameters (except for $\beta$) are fixed.
Once the different m-VAE models are trained, we encode the test set of MNIST (\cref{fig:vae1}\,C) to get MNIST-Nd and analyze its embeddings.
A (ten-fold) cross-validated random forest classifier \citep{breiman2001random} achieves comparable classification accuracy across dimensions ($\sim 90 \%$, \cref{fig:vae1}\,D).
Similiar accuracies suggest that our embeddings are indeed comparably separable across different dimensions, albeit t-SNE embeddings in higher dimensional datasets look somewhat more condensed (\cref{fig:vae1}\,E).
The accuracies are not close to state-of-the-art classifiers by design: realistic biological datasets are imperfect and our datasets share this property. 
\begin{figure}[t]
\floatconts
  {fig:fig2}
  {\vspace{-18pt}\caption{
  \textbf{A}: t-SNE colored with DISCO scores. 
  \textbf{B-C}: Density peaks for different local radii.
  Outliers in the upper-right are the suggested density peaks center points.
  }\label{fig:tsne}
  }{\includegraphics[width=\linewidth]{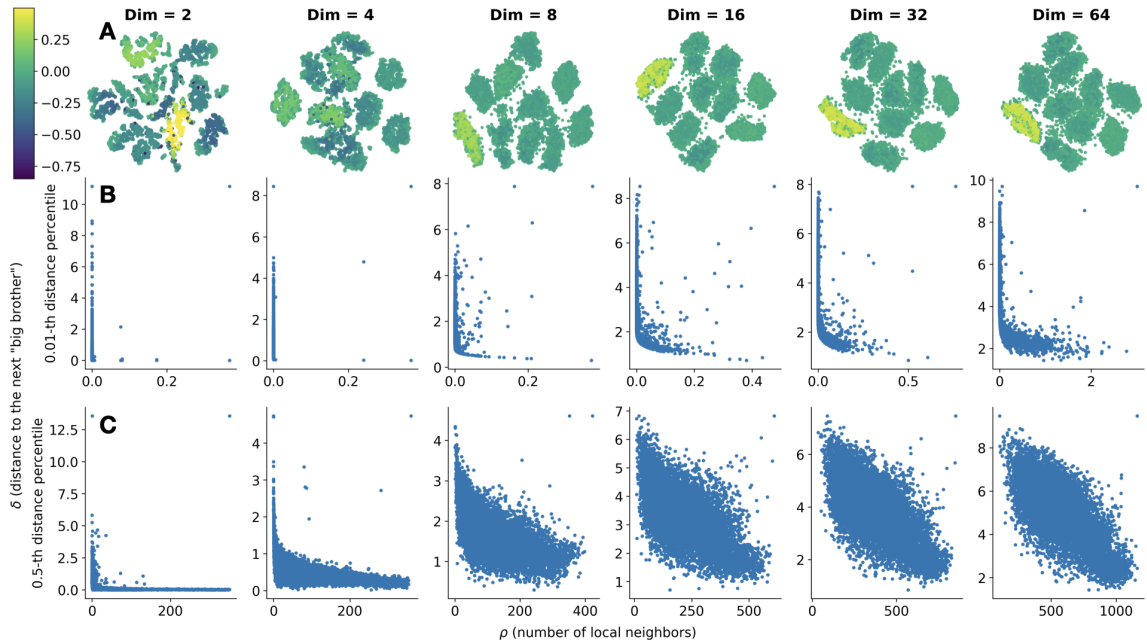}}
\end{figure}

\paragraph{Density modes and cluster overlap.}
As many real-world datasets have overlapping density between clusters, we want to ensure our toy datasets has it as well.
To check this, we estimated DISCO scores \citep{disco} (Fig.~2A) and search for density peaks (\cite{rodriguez2014clustering}) (Fig.~2B,\,C).
DISCO scores are bounded between $-1$ and 1, where negative values imply points being better connected to a different cluster than to the assigned one. 
The majority of points of our embeddings are scored around zero or below, suggesting a noticeable density overlap.
%
%
%
An alternative analysis is to count the density peaks. 
We follow \citet{rodriguez2014clustering} and measure density through the number of neighbors $\rho$ within a radius $r$ hypersphere (using an exponential kernel, so values are not integers).
Then for each point we compute $\delta$, the distance to the closest point with more neighbors. 
This way, the outliers in the upper-right corner of the $\rho$-$\delta$ diagram are natural cluster centers as they have many local neighbors and are far away from other points with more local neighbors.
The number of clear density modes is smaller than ten for all plots, indicating overlapping clusters.
As distances in high-dimensional space tend to be more uniformly distributed, smaller thresholds reveal more density peaks in higher dimensions. We used different thresholds to ensure same qualitative results (\cref{fig:tsne}\,B--C, \cref{apd:fast_search}).


\section{Evaluating clustering methods on MNIST-Nd}

Next, we use MNIST-Nd to test the performance and robustness of different clustering algorithms.
We choose $k$-means as an example for distance-based clustering, Gaussian and $t$-distribution mixture modelling (GMM and TMM) as density-based, and Leiden clustering as a graph-based clustering method.
For evaluation, we use the adjusted rand index (ARI) \citep{hubert1985comparing}, which measures the pairwise similarity of two cluster assignments. 
It is one when two partitions match exactly up to global label permutations. 
It is zero when the agreement between the two partitions is consistent with random assignments and negative when consistency is systematically below chance \citep{chacon2023minimum}.

First, we compare the clustering performance by calculating the ARI between cluster predictions and ground truth labels across ten random seeds.
The performance of the clustering algorithms decreases with growing dimensions except for Leiden clustering which does not seem to be affected as much (\cref{fig:robustnesses}\,A).
Second, we evaluate the stability of methods by computing the ARI between all pairs of partitions from the previous step.
While Leiden clustering remains stable, the ARI decreases with dimensions for the centroid-based methods (\cref{fig:robustnesses}\,B).
Third, we measure robustness to data perturbations
using bootstrapping.
We create three datasets, each containing 60\% of the original data. Of this, 40\% is shared across all three datasets for ARI estimation, while 20\% is unique to each dataset (\cref{fig:robustnesses}\,C). The datasets split is consistent across dimensions.
GMMs and $k$-means show the biggest decay along dimensions, while TMM clustering appeared to be more robust.
This is the only experiment where we see a decrease of Leiden ARIs for higher dimensions.
For all methods ARI values decline with higher dimensions because more points are needed to confidently estimate distances and densities in high-dimensional space.
\begin{figure}[t]
\floatconts
  {fig:robustnesses}
  {\vspace{-18pt}\caption{
  \textbf{A}: ARI with ground truth 
  shows clustering performances.
  \textbf{B}: ARI across seeds measures the stability of the clustering methods for different initializations.
  \textbf{C}: ARI for bootstrapped datasets shows clustering robustness for data perturbations.
  }
  }
  {\includegraphics[width=\linewidth]{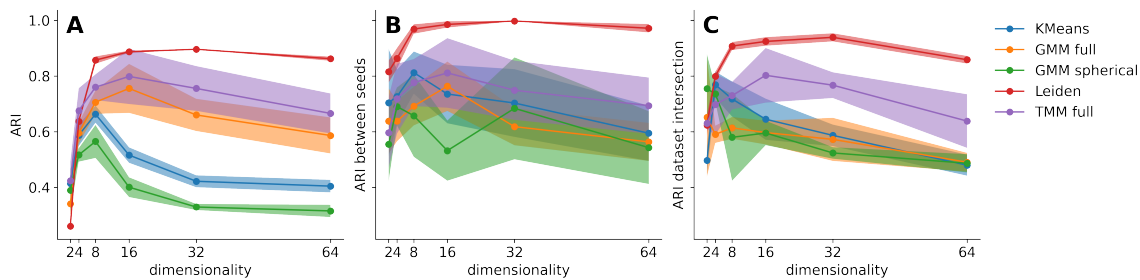}}
\end{figure}

\section{Conclusions and Limitations}
\label{sec:discussion}
We propose a framework to generate realistic noisy multidimensional datasets to isolate the impact of dimensionality on clustering performance. 
Benchmarking on MNIST embeddings shows that Leiden clustering outperforms other methods in higher dimensions, though this result needs validation on additional datasets.
Future work may include generating embeddings for other datasets to explore varying internal structures and validate these findings.
\newpage
\label{sec:cite}
\acks{
We thank Kenneth Harris, Ayush Paliwal and Paul Wollenhaupt for insightful discussions.
\\
Computing time was made available on the high-performance computers HLRN-IV at GWDG at the NHR Center NHR@Göttingen. 
The center is jointly supported by the Federal Ministry of Education and Research and the state governments participating in the NHR (www.nhr-verein.de/unsere-partner).
This project has received funding from the European Research Council (ERC) under the European Union’s Horizon Europe research and innovation programme (Grant agreement No. 101041669).
}

\bibliography{pmlr-sample}
\newpage

\appendix

\section{m-VAE and its Losses}\label{appendix:losses}
The m-VAE model has two main components: an encoder and a decoder. 
The encoder itself has two parts.
The first estimates the probability $p(y|x)$, where $x$ is the input and $y$ is the 'class label'. 
To achieve it, the it assigns a point to a mixture component using the Gumbel Softmax layer \citep{jang2016categorical}, which enables differentiable sampling from a categorical distribution without ground truth labels. 
It computes the log probabilities of all the classes in the distribution and adds them to noise from the Gumbel distribution, similar to the reparametrization trick in a classic VAE \citep{vae}.
The class with the highest value is treated as a one-hot label.
The second part concatenates this pseudo-label $y$ with the input $x$ to estimate the latent variable $z$ from a Gaussian prior, $p(z|x, y)$. The decoder then reconstructs the input from $z$, similar to a standard VAE.
The loss includes KL and reconstruction losses are same as for $\beta$-VAE, and the categorical loss, which is the entropy between logits and probabilities from the Gumbel Softmax layer. 
Note that the ground truth labels are not used anywhere during learning.


\begin{figure}[ht]
\floatconts
  {fig:app_losses}
  {\caption{
  \textbf{A:} Reconstruction loss.
  \textbf{B:} KL loss. We assumed Gaussian distribution.
  \\\textbf{C:} Classification Loss. It seems like latent dimensions of 2 and 4 are too shallow to separate the groups linearly.
}
  }{\includegraphics[width=\linewidth]{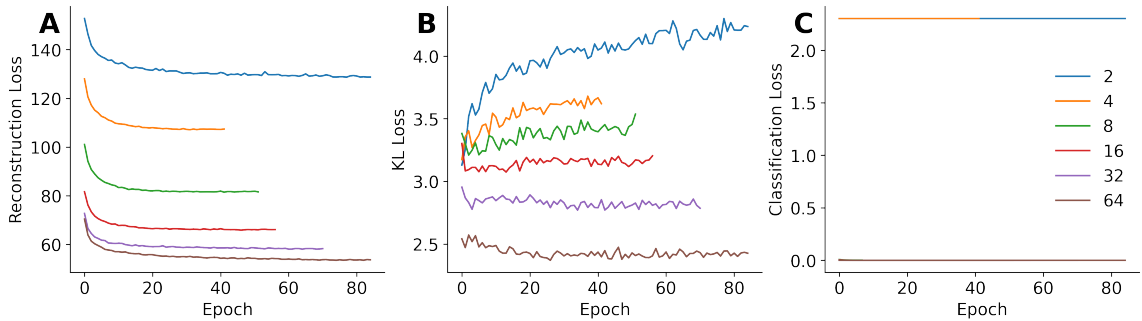}}
\end{figure}
\section{ARI explained}
To estimate clustering robustness, we use adjusted rand index (ARI) \citep{hubert1985comparing}, which measures the pairwise similarity of two cluster assignments $X$ and $Y$.
\begin{equation}
    ARI =\displaystyle \frac {\left.\sum _{ij}{\binom{n_{ij}}{2}}-\left[\sum _{i}\binom{a_i}{2}\sum _{j}\binom{b_j}{2}\right]\right/{\binom{n}{2}}}{\left.{\frac {1}{2}}\left[\sum _{i}{\binom{a_i}{2}}+\sum _{j}{\binom{b_j}{2}}\right]-\left[\sum _{i}{\binom{a_i}{2}}\sum _{j}{\binom{b_j}{2}}\right]\right/{\binom{n}{2}}} \ .
\end{equation}
Here, $a_{i}$ is the number of data points in cluster $i$ of partition $X$, $b_{j}$ is the number of data points in cluster $j$ of partition $Y$, $n_{ij}$ is the number of data points in clusters $i$ and $j$ of partition $X$ and $Y$, respectively, and $n$ is the total number of data points. 

ARI is an 
cluster validation index, which computes a score for a pair of partitions.
It checks for all pairs of points whether they are grouped together (end up in the same cluster) in both partitions.
If they are, it is counted as `agreement' while otherwise (the pair is in the same cluster in one partition and in different clusters in the other) the two partitions disagree.
It computes a score that reflects the proportion of agreements that are not due to random chance, meaning it accounts for the fact that some agreements would occur randomly.
A score of 1 means perfect agreement (i.e. the clusterings are equivalent up to change of labels), 0 indicates that there are not more agreements between the two partitions than there would be between two random partitions.

\section{Other clustering evaluation metrics}
\label{apd:cluster_metrics}

There are other metrics than ARI to evaluate clustering partitions. 
ARI is a metric between two clustering partitions as it evaluates if pairs of points end up in the same groups across partitions.
\emph{Fowlkes-Mallows} also compares two clustering partitions but, in contrast to ARI, without adjusting for chance.
The score is the geometric mean of the precision and recall across the whole dataset (true positive: pairs of points grouped together in both partitions).
The other three set-based metrics are homogeneity, completeness, and v-measure.
These three scores are asymmetric, one partition is being evaluated while the other provides (pseudo)labels.
Ideally, this `other' partition would be the ground-truth assignment.
A clustering result satisfies homogeneity if all of its clusters contain only data points which are members of a single class.
The score is achieved by computing the fraction of points with the `correct' label, averaged over all clusters.
A clustering result satisfies completeness if all the data points that are members of a given class are elements of the same cluster.
Both metrics are not symmetric: switching ground-truth and predicted labels for completeness will return the homogeneity, and vice versa. 
V-measure is equivalent to the normalized mutual information (NMI) and it's the harmonic mean of completeness and homogeneity.

We see that the trends and line order in \cref{fig:clustscores} is consistent with ARI across all the metrics, meaning that the results based on ARI are compatible with those other cluster evaluation metrics.

\begin{figure}[ht]
\floatconts
  {fig:clustscores}
  {\caption{
  \textbf{A:} Fowlkes-Mallows score.
  \textbf{B:} Homogeneity.
  \textbf{C:} Completeness.
  \textbf{D:} NMI.}
  }{\includegraphics[width=\linewidth]{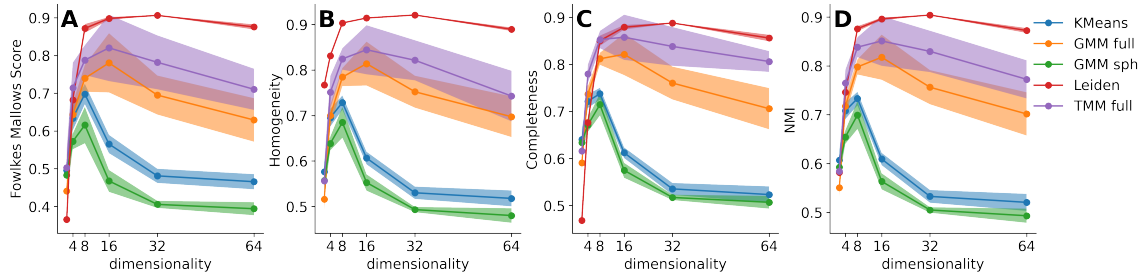}}
\end{figure}

\section{Cluster Validity Indices on MNIST-Nd}
\label{apd:cvi}

\subsection{Internal Cluster Validation}

In the main paper we provided DISCO scores which are effectively Silhouette scores adapted to a density-based setting.
In contrast to Silhouette which strictly favors ball-shaped clusters, density-based cluster validation indices such as DBCV \citep{dbcv}, DCSI \citep{dcsi}, and DISCO only look at the `gap' between clusters but work for clusters of arbitrary shapes.

We chose to use DISCO scores in \cref{fig:tsne} as it outputs a point-wise score, allowing us to visualize that essentially all clusters are slightly overlapping, as we desired.

For comparison, we provide (overall) scores computed with different density-based cluster validation indices in \cref{tab:other_cvis} and S\_Dbw \citep{sdbw} as a centroid-based CVI ($\downarrow$ indicates that for S\_Dbw lower values are better).
From the table, we can see that both DISCO and DBCV evaluate the clustering as `bad' (DBCV is more prone to output -1 compared to DISCO).
While DISCO and S\_Dbw consider all of the embeddings equally bad, DBCV and DCSI clearly prefer the 8-64 dimensional embeddings, even though they do not look more easily separable in t-SNE (see  \cref{fig:tsne}).
Overall, the values indicate that there is still a significant or large amount of overlap between clusters as desired.

\begin{table}[h]
    \centering
    \begin{tabular}{ccccc}
    \toprule
        & DISCO & DBCV & DCSI & S\_Dbw ($\downarrow$)\\
        \midrule
        2d  & -0.05 & -1    & 0.16 & 0.87\\
        4d  & -0.10 & -0.92 & 0.35 & 0.78\\
        8d  & -0.02 & -0.77 & 0.48 & 0.68\\
        16d & -0.01 & -0.65 & 0.46 & 0.81\\
        32d & -0.02 & -0.63 & 0.46 & 0.86\\
        64d & -0.02 & -0.56 & 0.44 & 0.88\\
        \bottomrule
    \end{tabular}
    \caption{CVI scores across dimensionalities. There is significant overlap between the clusters.}
    \label{tab:other_cvis}
\end{table}

\section{Fast search density peak analysis}
\label{apd:fast_search}
\subsection{Different thresholds}
The main hyperparameter of fast density peak search algorithm \citep{rodriguez2014clustering}
is radius $r$ for local density estimation.
As distances become more uniform in the higher dimensions, it is crucial to scale the radius with it.
To avoid being biased for the choice of $r$ hyperparameter, we performed this analysis using  $[0.01, 0.1, 0.3, 0.5, 1, 2, 3, 5]$-th distances percentiles as $r$. 
As expected, smaller distances reveal more density peaks for higher dimensions but generally we see below ten clear density modes within a range of thresholds.
\begin{figure}[ht]
\floatconts
  {fig:rhodeltabelow1}
  {\caption{$\rho$-$\delta$ plots for different threshold values. Thresholds are percentiles from the distribution of all distances. In this plot the percentiles below one percent. Pth stays for percentile. The clear separated points are natural cluster centers.}
  }{\includegraphics[width=\linewidth]{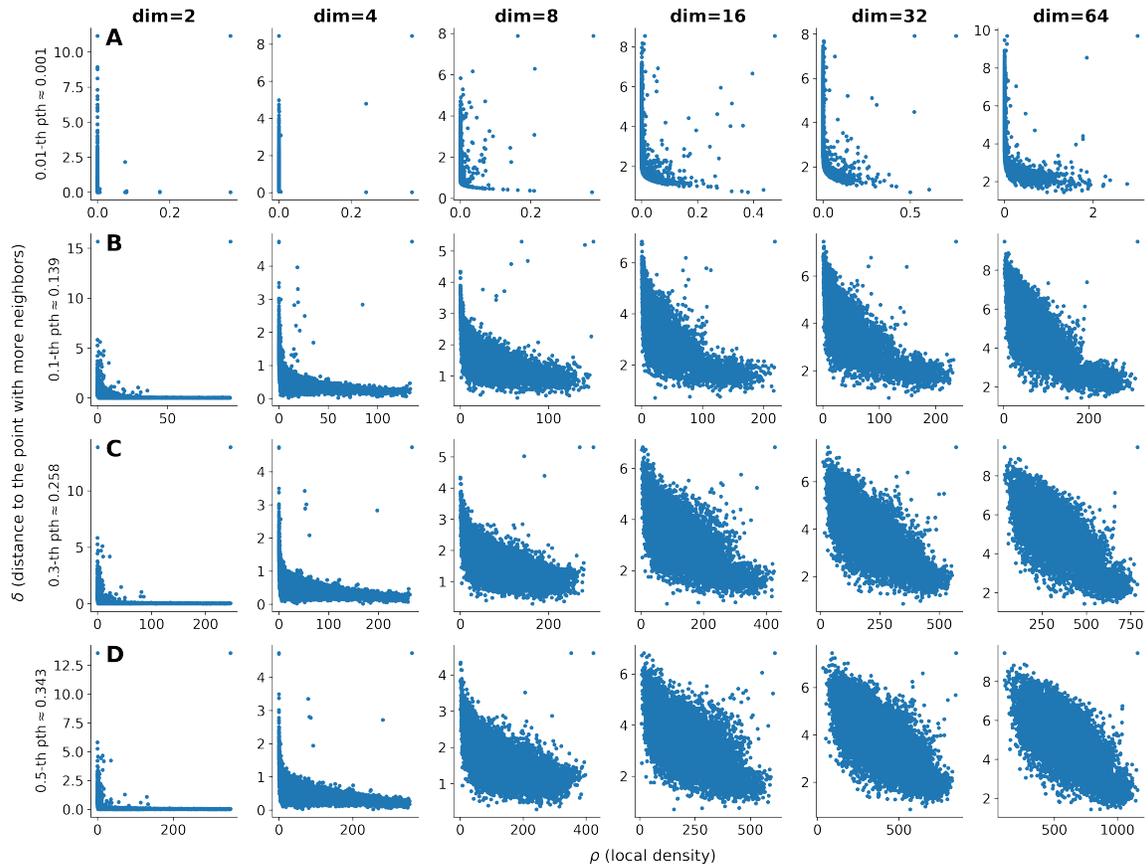}}
\end{figure}

\begin{figure}[ht]
\floatconts
  {fig:rhodeltaabove1}
  {\caption{Same as \cref{fig:rhodeltabelow1} but percentiles are now above one percent.}
  }{\includegraphics[width=\linewidth]{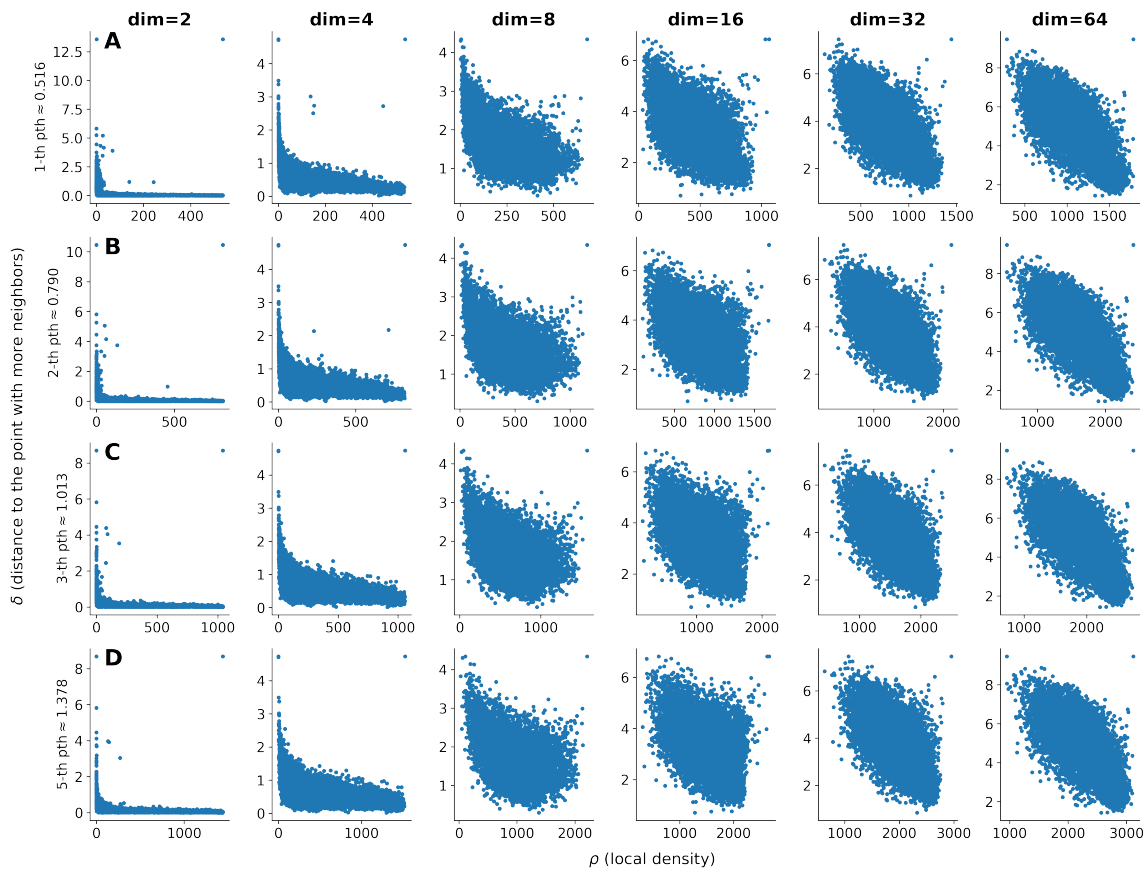}}
\end{figure}
\subsection{Gamma cluster centers selection}
Following \cite{rodriguez2014clustering} one could use a product of local density and distances $\gamma = \rho \cdot \delta$ to the neighbor with bigger amount of neighbors to select the clusters centers. 
If the number of clusters is defined as $n$, one would just take $n$ points with the biggest $\gamma$.
Otherwise, order points by $\gamma$ decreasing and see how many of the first points have substantially big gaps between each other and select these as cluster centers. 
Please note that this strategy works nicely for balanced clusters, which is the case for us, otherwise, other strategies should be applied \cite{sieranoja2019fast}.
Below we provide plots for the points with 30 biggest $\gamma$ values for different thresholds.
\begin{figure}[t]
\floatconts
  {fig:gammabelow1}
  {\caption{$\gamma$ values 
  for different threshold values. Thresholds are percentiles from the distribution of all distances. In this plot the percentiles below one percent. Pth stays for percentile. The clear separated points are natural cluster centers.}
  }{\includegraphics[width=\linewidth]{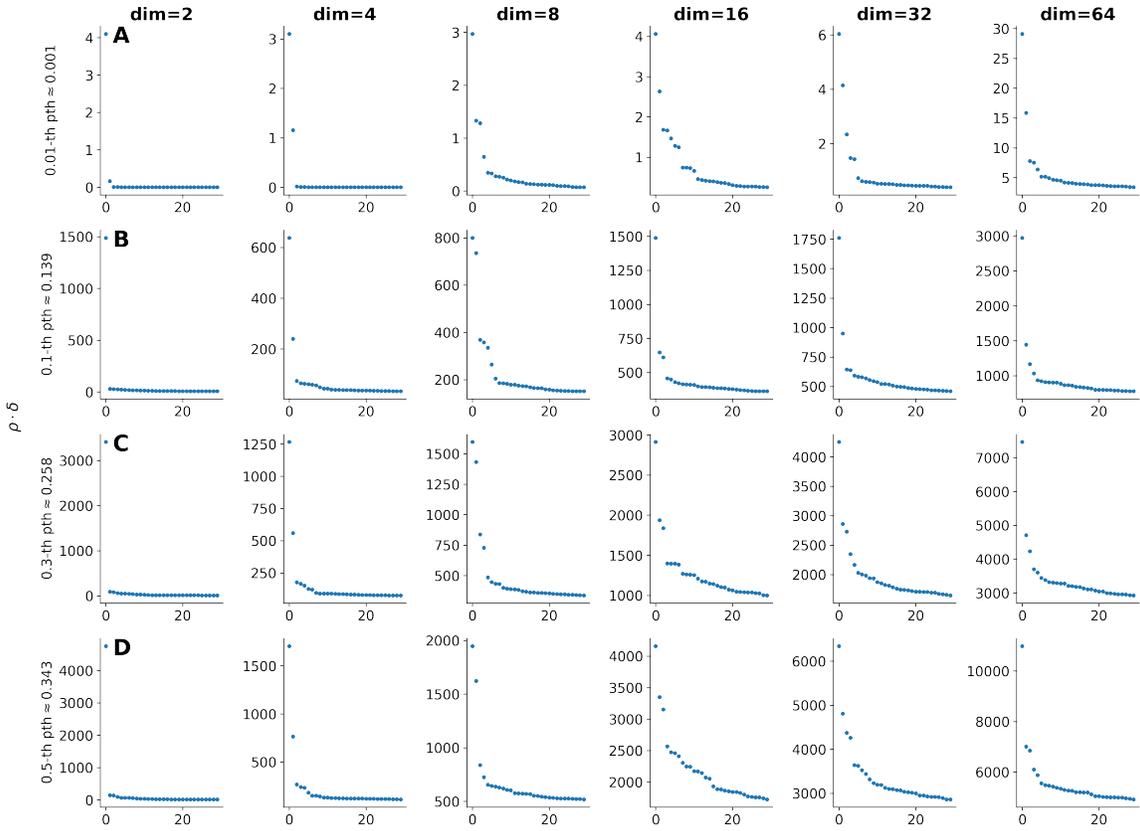}}
\end{figure}

\begin{figure}[t]
\floatconts
  {fig:gammaabove1}
  {\caption{Same as \cref{fig:gammabelow1} but percentiles are now above one percent.}
  }
  {\includegraphics[width=\linewidth]{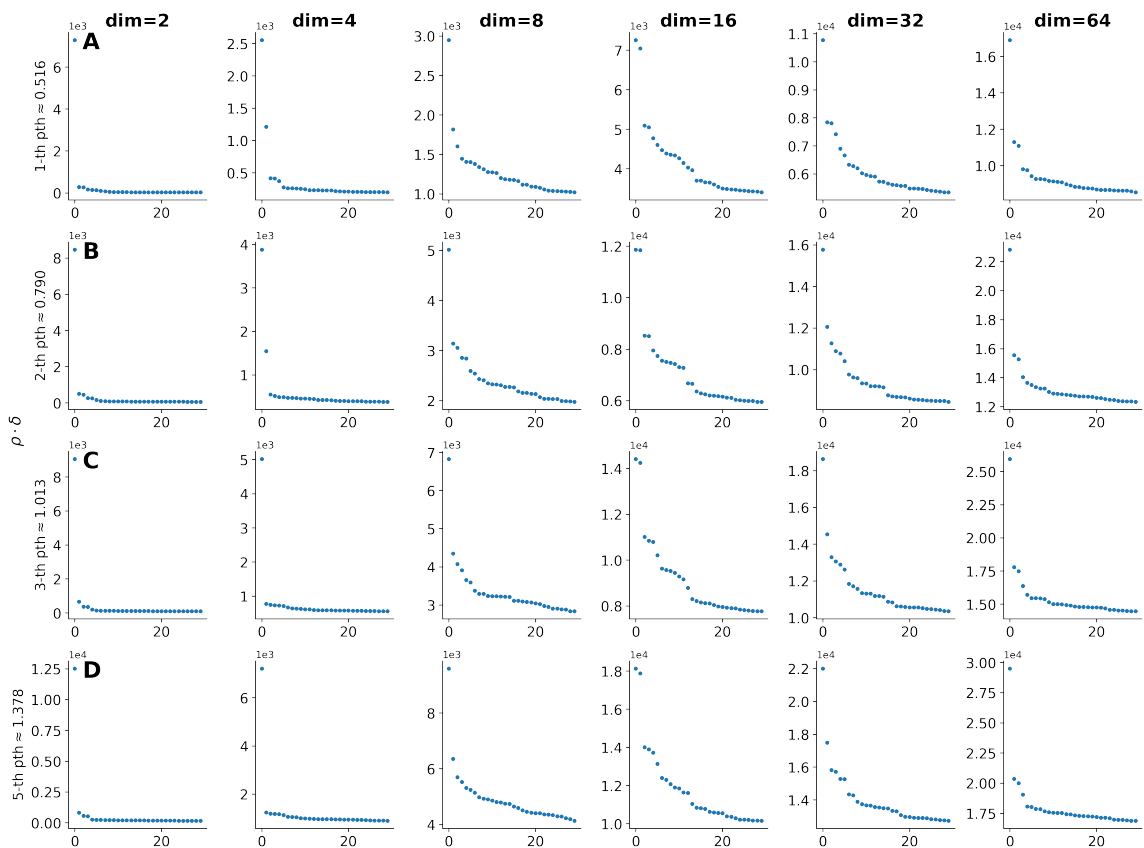}}
\end{figure}
\section{Code sources acknowledgement}
\begin{itemize}
    \item We adjusted the PyTorch version of the following repo to train a m-VAE \\\url{https://github.com/jariasf/GMVAE.git}
    \item We also adjusted code from here for early stopping \\\url{https://github.com/Bjarten/early-stopping-pytorch}
    \item For TMM we used the implementation from \url{https://github.com/jlparki/mix_T}. 
    \item We used Leiden cluster implementation from `scanpy` package, \citep{wolf2018scanpy} \\\url{https://scanpy.readthedocs.io/en/stable/}, 
    \item K-means and GMM were used from the `scikit-learn` package \citep{sklearn_api}
\end{itemize}
\end{document}